\documentclass[11pt,a4paper]{article}
\usepackage[hyperref]{emnlp2018}
\usepackage{times}
\usepackage{latexsym}
\usepackage{anysize}
\usepackage{graphicx}
\usepackage{mathtools}
\usepackage{amsmath}
\usepackage{bm}
\usepackage{enumerate}
\usepackage{multirow}
\usepackage{algpseudocode}
\usepackage{algorithmicx}
\usepackage{subfig}
\usepackage{caption}
\usepackage{float}
\usepackage{url}
\usepackage{todonotes}
\usepackage{arydshln}

\aclfinalcopy 
\def\aclpaperid{***} 

\title{Neural Character-based Composition Models for Abuse Detection}

\author{
  Pushkar Mishra \\
  Dept. of CS \& Technology\\
  University of Cambridge\\
  United Kingdom\\
  {\tt pm576@alumni.cam.ac.uk} \\ \And
  \; \; \;Helen Yannakoudakis\\
  \; \; \;The ALTA Institute\\
  \; \; \;University of Cambridge\\
  \; \; \;United Kingdom\\
  \; \; \;{\tt hy260@cl.cam.ac.uk} \\ \And
  Ekaterina Shutova\\
  ILLC\\
  University of Amsterdam\\
  The Netherlands\\
  {\tt e.shutova@uva.nl}
}

\date{}

\begin{document}
\maketitle
\begin{abstract} 
The advent of social media in recent years has fed into some highly undesirable phenomena such as proliferation of offensive language, hate speech, sexist remarks, etc. on the Internet. In light of this, there have been several efforts to automate the detection and moderation of such abusive content. However, deliberate obfuscation of words by users to evade detection poses a serious challenge to the effectiveness of these efforts. The current state of the art approaches to abusive language detection, based on recurrent neural networks, do not explicitly address this problem and resort to a generic \textsc{oov} (out of vocabulary) embedding for unseen words. However, in using a single embedding for all unseen words we lose the ability to distinguish between obfuscated and non-obfuscated or rare words.
In this paper, we address this problem by designing a model that can compose embeddings for unseen words. We experimentally demonstrate that our approach significantly advances the current state of the art in abuse detection on datasets from two different domains, namely Twitter and Wikipedia talk page.
\end{abstract}

\section{Introduction}
\textit{Pew Research Center} has recently uncovered several disturbing trends in communications on the Internet. As per their report \cite{pew}, 40\% of adult Internet users have personally experienced harassment online, and 60\% have witnessed the use of offensive names and expletives. Expectedly, the majority (66\%) of those who have personally faced harassment have had their most recent incident occur on a social networking website or app. While most of these websites and apps provide ways of flagging offensive and hateful content, only 8.8\% of the victims have actually considered using such provisions.

Two conclusions can be drawn from these statistics: (i) \textit{abuse} (a term we use henceforth to collectively refer to toxic language, hate speech, etc.) is prevalent in social media, and (ii) passive and/or manual techniques for curbing its propagation (such as flagging) are neither effective nor easily scalable \cite{Pavlopoulos:17}. Consequently, the efforts to automate the detection and moderation of such content have been gaining popularity \cite{c53cecce142c48628b3883d13155261c,Wulczyn:2017:EMP:3038912.3052591}.

In their work, Nobata et al. \shortcite{Nobata:2016:ALD:2872427.2883062} describe the task of achieving effective automation as an inherently difficult one due to several ingrained complexities; a prominent one they highlight is the deliberate structural obfuscation of words (for example, \textit{fcukk}, \textit{w0m3n}, \textit{banislam}, etc.) by users to evade detection. Simple spelling correction techniques and edit-distance procedures fail to provide information about such obfuscations because: (i) words may be excessively fudged (e.g., \textit{a55h0le}, \textit{n1gg3r}) or concatenated (e.g., \textit{stupidbitch}, \textit{feminismishate}), and (ii) they fail to take into account the fact that some character sequences like \textit{musl} and \textit{wom} are more frequent and more indicative of abuse than others \cite{c53cecce142c48628b3883d13155261c}.

Nobata et al. \shortcite{Nobata:2016:ALD:2872427.2883062} go on to show that simple character n-gram features prove to be highly promising for supervised classification approaches to abuse detection due to their robustness to spelling variations; however, they do not address obfuscations explicitly. Waseem and Hovy \shortcite{c53cecce142c48628b3883d13155261c} and Wulczyn et al. \shortcite{Wulczyn:2017:EMP:3038912.3052591} also use character n-grams to attain impressive results on their respective datasets. That said, the current state of the art methods do not exploit character-level information, but instead utilize recurrent neural network (\textsc{rnn}) models operating on word embeddings alone \cite{Pavlopoulos:17,Badjatiya:17}. Since the problem of deliberately noisy input is not explicitly accounted for, these approaches resort to the use of a generic \textsc{oov} (out of vocabulary) embedding for words not seen in the training phase. However, in using a single embedding for all unseen words, such approaches lose the ability to distinguish obfuscated words from non-obfuscated or rare ones. Recently, Mishra et al. \shortcite{mishra} and Qian et al. \shortcite{leveraging}, working with the same Twitter dataset as we do, reported that many of the misclassifications by their \textsc{rnn}-based methods happen due to intentional misspellings and/or rare words.

Our contributions are two-fold: first, we experimentally demonstrate that character n-gram features are complementary to the current state of the art \textsc{rnn} approaches to abusive language detection and can strengthen their performance. We then explicitly address the problem of deliberately noisy input by constructing a model that operates at the character level and learns to predict embeddings for unseen words. We show that the integration of this model with the character-enhanced \textsc{rnn} methods further advances the state of the art in abuse detection on three datasets from two different domains, namely Twitter and Wikipedia talk page. To the best of our knowledge, this is the first work to use character-based word composition models for abuse detection.

\section{Related Work}
Yin et al. \shortcite{Yin09detectionof} were among the first ones to apply supervised learning to the task of abuse detection. They worked with a linear support vector machine trained on local (e.g., n-grams), contextual (e.g., similarity of a post to its neighboring posts), and sentiment-based (e.g., presence of expletives) features to recognize posts involving harassment.

Djuric et al. \shortcite{Djuric:2015:HSD:2740908.2742760} worked with comments taken from the Yahoo Finance portal and demonstrated that distributional representations of comments learned using the \textit{paragraph2vec} framework \cite{DBLP:journals/corr/LeM14} can outperform simpler bag-of-words \textsc{bow} features under supervised classification settings for hate speech detection. Nobata et al. \shortcite{Nobata:2016:ALD:2872427.2883062} improved upon the results of Djuric et al. by training their classifier on an amalgamation of features derived from four different categories: linguistic (e.g., count of insult words), syntactic (e.g. part-of-speech \textsc{pos} tags), distributional semantic (e.g., word and comment embeddings) and n-gram based (e.g., word bi-grams). They noted that while the best results were obtained with all features combined, character n-grams had the highest impact on performance.

Waseem and Hovy \shortcite{c53cecce142c48628b3883d13155261c} utilized a logistic regression (\textsc{lr}) classifier to distinguish amongst racist, sexist, and clean tweets in a dataset of approximately $16k$ of them. They found that character n-grams coupled with gender information of users formed the optimal feature set for the task. On the other hand, geographic and word-length distribution features provided little to no improvement. Experimenting with the same dataset, Badjatiya et al. \shortcite{Badjatiya:17} improved on their results by training a gradient-boosted decision tree (\textsc{gbdt}) classifier on averaged word embeddings learnt using a long short-term memory (\textsc{lstm}) models initialized with random embeddings. Mishra et al. \shortcite{mishra} went on to incorporate community-based profiling features of users in their classification methods, which led to the state of the art performance on this dataset. 

Waseem \shortcite{zeerakW16-5618} studied the influence of annotators' knowledge on the task of hate speech detection. For this, they sampled $7k$ tweets from the same corpus as Waseem and Hovy \shortcite{c53cecce142c48628b3883d13155261c} and recruited expert and amateur annotators to annotate the tweets as \textit{racism}, \textit{sexism}, \textit{both} or \textit{neither}. Combining this dataset with that of Waseem and Hovy \shortcite{c53cecce142c48628b3883d13155261c}, Park et al. \shortcite{W17-3006} evaluated the efficacy of a 2-step classification process: they first used an \textsc{lr} classifier to separate abusive and non-abusive tweets, and then used another \textsc{lr} classifier to distinguish between the racist and sexist ones. They showed that this setup had comparable performance to a 1-step classification approach based on convolutional neural networks (\textsc{cnn}s) operating on word and character embeddings. 

Wulczyn et al. \shortcite{Wulczyn:2017:EMP:3038912.3052591} created three different datasets of comments collected from the English Wikipedia Talk page: one was annotated for personal attacks, another for toxicity, and the third for aggression. They achieved their best results with a multi-layered perceptron classifier trained on character n-gram features. Working with the personal attack and toxicity datasets, Pavlopoulos et al. \shortcite{Pavlopoulos:17} outperformed the methods of Wulczyn et al. by having a gated recurrent unit (\textsc{gru}) to model the comments as dense low-dimensional representations, followed by an \textsc{lr} layer to classify the comments based on those representations.

Davidson et al. \shortcite{davidson} produced a dataset of about $25k$ \textit{racist}, \textit{offensive} or \textit{clean} tweets. They evaluated several multi-class classifiers with the aim of discerning clean tweets from racist and offensive tweets, while simultaneously being able to distinguish between the racist and offensive ones. Their best model was an \textsc{lr} classifier trained using \textsc{tf--idf} and \textsc{pos} n-gram features coupled with features like count of hash tags and number of words.

\section{Datasets}
Following the proceedings of the \textit{1$^{st}$ Workshop on Abusive Language Online} \cite{waseem2017proceedings}, we use three datasets from two different domains.

\subsection{Twitter}
Waseem and Hovy \shortcite{c53cecce142c48628b3883d13155261c} prepared a dataset of $16,914$ tweets from a corpus of approximately $136k$ tweets retrieved over a period of two months. They bootstrapped their collection process with a search for commonly used slurs and expletives related to religious, sexual, gender and ethnic minorities. After having manually annotated $16,914$ of the tweets as \textit{racism}, \textit{sexism} or \textit{neither}, they asked an expert to review their annotations in order to mitigate against any biases. The inter-annotator agreement was reported at $\kappa=0.84$, with further insight that $85\%$ of all the disagreements occurred in the \textit{sexism} class alone.

The authors released the dataset as a list of $16,907$ tweet \textsc{id}s and their corresponding annotations. We could only retrieve $16,202$ of the tweets with python's \textit{Tweepy} library since some of them have been deleted or their visibility has been limited. Of the ones retrieved, 1,939 (12\%) are \textit{racism}, 3,148 (19.4\%) are \textit{sexism}, and the remaining 11,115 (68.6\%) are \textit{neither}; the original dataset has a similar distribution, i.e., 11.7\% \textit{racism}, 20.0\% \textit{sexism}, and 68.3\% \textit{neither}.

\subsection{Wikipedia talk page}
Wulczyn et al. \shortcite{Wulczyn:2017:EMP:3038912.3052591} extracted approximately $63M$ talk page comments from a public dump of the full history of English Wikipedia released in January 2016. From this corpus, they randomly sampled comments to form three datasets on personal attack, toxicity and aggression, and engaged workers from \textit{CrowdFlower} to annotate them. Noting that the datasets were highly skewed towards the non-abusive classes, the authors over-sampled comments from banned users to attain a more uniform distribution.

In this work, we utilize the toxicity and personal attack datasets, henceforth referred to as \textsc{w-tox} and \textsc{w-att} respectively. Each comment in both of these datasets was annotated by at least 10 workers. We use the majority annotation of each comment to resolve its gold label: if a comment is deemed toxic (alternatively, attacking) by more than half of the annotators, we label it as \textit{abusive}; otherwise, as \textit{non-abusive}. 13,590 (11.7\%) of the 115,864 comments in \textsc{w-att} and 15,362 (9.6\%) of the 159,686 comments in \textsc{w-tox} are abusive. Wikipedia comments, with an average length of 25 tokens, are considerably longer than the tweets which have an average length of 8.

\section{Methods}
We experiment with ten different methods, eight of which have an \textsc{rnn} operating on word embeddings. Six of these eight also include character n-gram features, and four further integrate our word composition model. The remaining two comprise an \textsc{rnn} that works directly on character inputs.

\vspace{1 mm}
\noindent
\textbf{Hidden-state (\textsc{hs}).} As our first baseline, we adopt the ``\textsc{rnn}'' method of Pavlopoulos et al. \shortcite{Pavlopoulos:17} since it produces state of the art results on the Wikipedia datasets. Given a text formed of a sequence $w_1$, $\dots$, $w_n$ of words (represented by $d$-dimensional word embeddings), the method utilizes a 1-layer \textsc{gru} to encode the words into hidden states $h_1$, $\dots$, $h_n$. This is followed by an \textsc{lr} layer that classifies the text based on the last hidden state $h_n$. We modify the authors' original architecture in two minor ways: we extend the 1-layer \textsc{gru} to a 2-layer \textsc{gru} and use softmax as the activation in the \textsc{lr} layer instead of sigmoid.\footnote{We also experimented with 1-layer \textsc{gru}/\textsc{lstm} and 1/2-layer bi-directional \textsc{gru}s/\textsc{lstm}s but the performance only worsened or showed no gains; using sigmoid instead of softmax did not have any noteworthy effects on the results either.}

Following Pavlopoulos et al., we initialize the word embeddings to \textsc{gl}o\textsc{v}e vectors \cite{Pennington:14}. In all our methods, words not present in the \textsc{gl}o\textsc{v}e set are randomly initialized in the range $\pm 0.05$, indicating the lack of semantic information. By not mapping these words to a single random embedding, we mitigate against the errors that may arise due to their conflation \cite{Madhyastha:15}. A special \textsc{oov} (out of vocabulary) token is also initialized in the same range. All the embeddings are updated during training, allowing for some of the randomly-initialized ones to get task-tuned \cite{D14-1181}; the ones that do not get tuned lie closely clustered around the \textsc{oov} token to which unseen words in the test set are mapped.

\vspace{1 mm}
\noindent
\textbf{Word-sum (\textsc{ws}).} The ``\textsc{lstm}+\textsc{gl}o\textsc{v}e+\textsc{gbdt}" method of Badjatiya et al. \shortcite{Badjatiya:17} constitutes our second baseline. The authors first employ an \textsc{lstm} to task-tune \textsc{gl}o\textsc{v}e-initialized word embeddings by propagating error back from an \textsc{lr} layer. They then train a gradient-boosted decision tree (\textsc{gbdt}) classifier to classify texts based on the average of the constituent word embeddings.\footnote{In their work, the authors report that initializing embeddings randomly rather than with \textsc{gl}o\textsc{v}e yields state of the art performance on the Twitter dataset that we are using. However, we found the opposite when performing 10-fold stratified cross-validation (\textsc{cv}). A possible explanation of this lies in the authors' decision to not use stratification, which for such a highly imbalanced dataset can lead to unexpected outcomes \cite{Forman:10}. Furthermore, the authors train their \textsc{lstm} on the entire dataset including the test part without any early stopping criterion; this facilitates over-fitting of the randomly-initialized embeddings.} We make two minor modifications to the original method: we utilize a 2-layer \textsc{gru}\footnote{The deeper 2-layer \textsc{gru} slightly improves performance.} instead of the \textsc{lstm} to tune the embeddings, and we train the \textsc{gbdt} classifier on the \textsc{l}$_2$-normalized sum of the embeddings instead of their average.\footnote{\textsc{l}$_2$-normalized sum ensures uniformity of range across the feature set in all our methods; \textsc{gbdt}, being a tree based model, is not affected by the choice of monotonic function.}

\vspace{1 mm}
\noindent
\textbf{Hidden-state + char n-grams (\textsc{hs + cng}).} Here we extend the \textit{hidden-state} baseline: we train the 2-layer \textsc{gru} architecture as before, but now concatenate its last hidden state $h_n$ with \textsc{l}$_2$-normalized character n-gram counts to train a \textsc{gbdt} classifier.

\vspace{1 mm}
\noindent
\textbf{Augmented hidden-state + char n-grams (\textsc{augmented hs + cng}).} In the above methods, unseen words in the test set are simply mapped to the \textsc{oov} token since we do not have a way of obtaining any semantic information about them. However, this is undesirable since racial slurs and expletives are often deliberately fudged by users to prevent detection. In using a single embedding for all unseen words, we lose the ability to distinguish such obfuscations from other non-obfuscated or rare words. Taking inspiration from the effectiveness of character-level features in abuse detection, we address this issue by having a character-based word composition model that can compose embeddings for unseen words in the test set \cite{pinter}. We then augment the \textit{hidden-state + char n-grams} method with it.

\begin{figure*}[t!]
\centering
\subfloat[]{
\includegraphics[width=8.75cm]{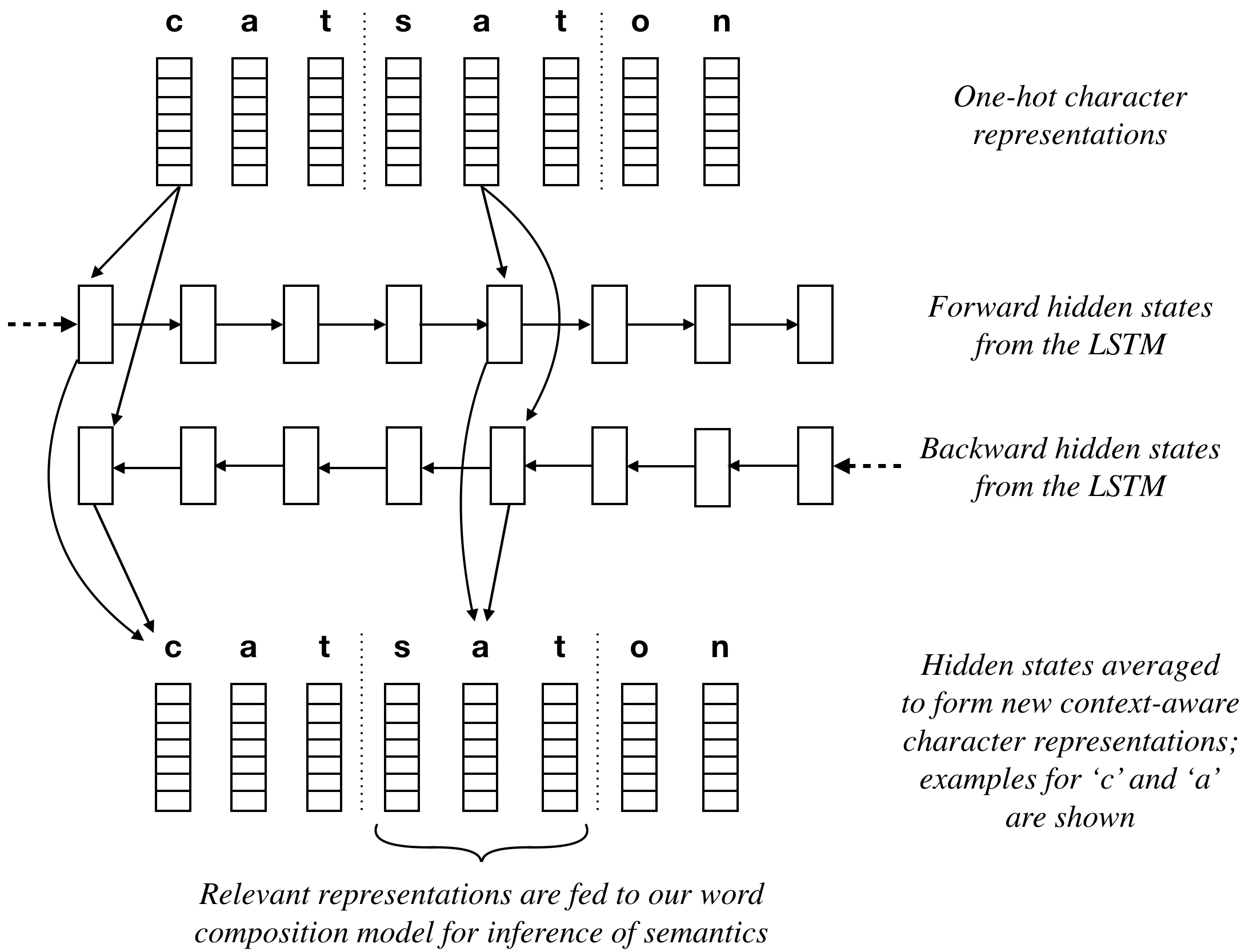}
}
\qquad\qquad\quad
\subfloat[]{
\label{word_model}
\includegraphics[width=3.83cm]{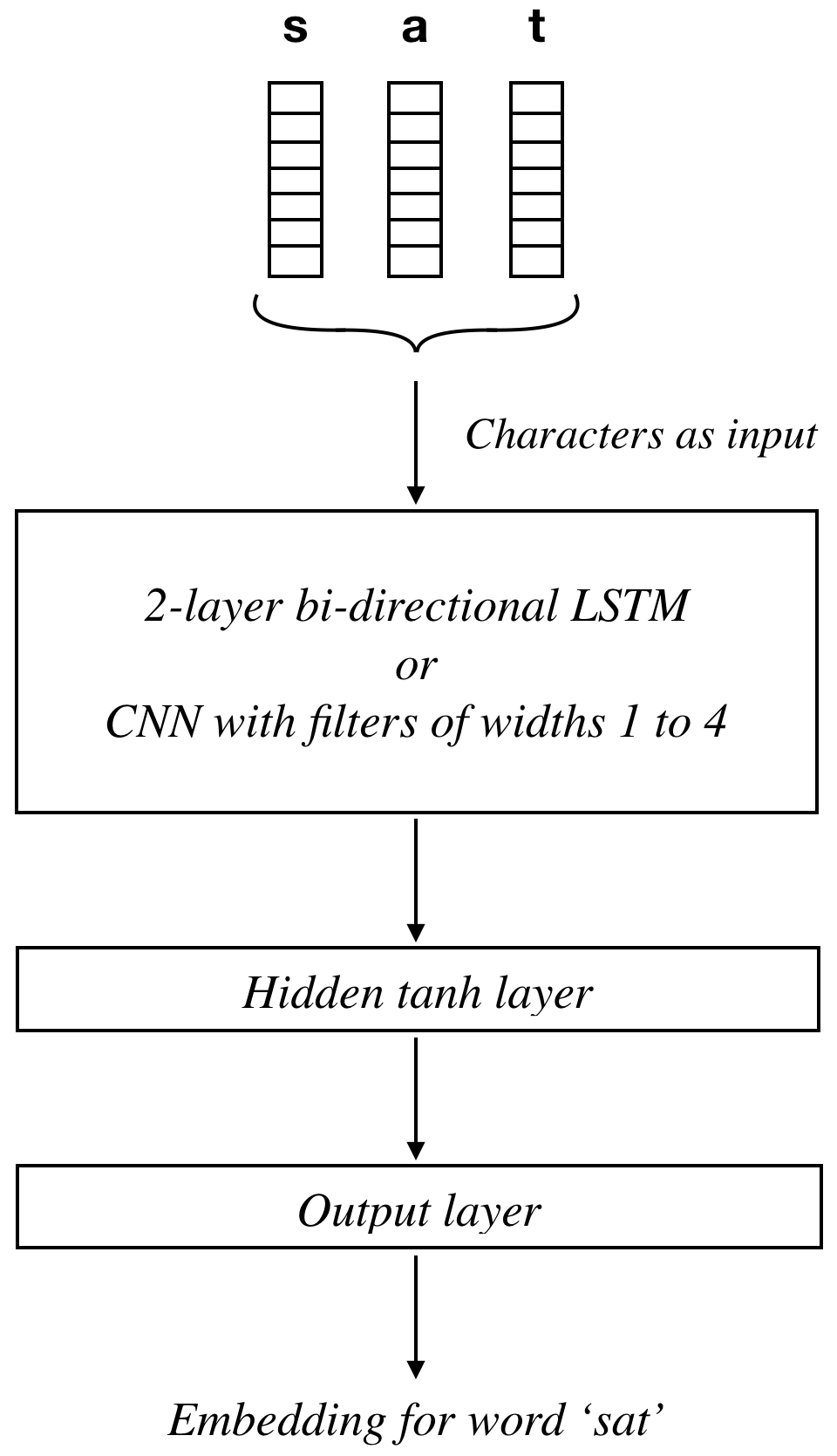}
}
\caption{Context-aware approach to word composition. The figure on the left shows how the encoder extracts context-aware representations of characters in the phrase ``\textit{cat sat on}" from their one-hot representations. The dotted lines denote the space character $_\sqcup$ which demarcates word boundaries. Semantics of an unseen word, e.g., \textit{sat}, can then be inferred by our word composition model shown on the right.}
\label{context}
\end{figure*}

Specifically, our model (Figure \ref{word_model}) comprises a 2-layer bi-directional \textsc{lstm}, followed by a hidden layer with \textit{tanh} non-linearity and an output layer at the end. The model takes as input a sequence $c_1$, $\dots$, $c_k$ of characters, represented as one-hot vectors, from a fixed vocabulary (i.e., lowercase English alphabet and digits) and outputs a $d$-dimensional embedding for the word `$c_1\dots c_k$'. Bi-directionality of the \textsc{lstm} allows for the semantics of both the prefix and the suffix (last hidden forward and backward state) of the input word to be captured, which are then combined to form the hidden state for the input word. The model is trained by minimizing the mean squared error (\textsc{mse}) between the embeddings that it produces and the task-tuned embeddings of words in the training set. This ensures that newly composed embeddings are endowed with characteristics from both the \textsc{gl}o\textsc{v}e space as well as the task-tuning process. While approaches like that of Bojanowski et al. \shortcite{fasttext} can also compose embeddings for unseen words, they cannot endow the newly composed embeddings with characteristics from the task-tuning process; this may constitute a significant drawback \cite{D14-1181}.

During the training of our character-based word composition model, to emphasize frequent words, we feed a word as many times as it appears in the training corpus. We note that a 1-layer \textsc{cnn} with global max-pooling in place of the 2-layer \textsc{lstm} provides comparable performance while requiring significantly less time to train. This is expected since words are not very long sequences, and the filters of the \textsc{cnn} are able to capture the different character n-grams within them.

\vspace{1 mm}
\noindent
\textbf{Context hidden-state + char n-grams (\textsc{context hs + cng}).} In the \textit{augmented hidden-state + char n-grams} method, the word composition model infers semantics of unseen words solely on the basis of the characters in them. However, for many words, semantic inference and sense disambiguation require context, i.e., knowledge of character sequences in the vicinity. An example is the word \textit{cnt} that has different meanings in the sentences ``\textit{I cnt undrstand this!}" and ``\textit{You feminist cnt!}", i.e., \textit{cannot} in the former and the sexist slur \textit{cunt} in the latter. Yet another example is an obfuscation like `'\textit{You mot otherf uc ker!} where the expletive \textit{motherfucker} cannot be properly inferred from any fragment without the knowledge of surrounding character sequences.

To address this, we develop \textit{context-aware representations} for characters as inputs to our character-based word composition model instead of one-hot representations.\footnote{We also experimented with word-level context but did not get any significant improvements. We believe this is due to higher variance at word level than at the character level.} We introduce an encoder architecture to produce the context-aware representations. Specifically, given a text formed of a sequence $w_1$, $\dots$, $w_n$ of words, the encoder takes as input one-hot representations of the characters $c_1$, $\dots$, $c_k$ within the concatenated sequence `$w_1{_\sqcup}\dots{_\sqcup}w_n$', where ${_\sqcup}$ denotes the space character. This input is passed through a bi-directional \textsc{lstm} that produces hidden states $h_1$, $\dots$, $h_k$, one for every character. Each hidden state, referred to as context-aware character representation, is the average of its designated forward and backward states; hence, it captures both the preceding as well as the following contexts of the character it corresponds to. Figure \ref{context} illustrates how the context-aware representations are extracted and used for inference by our character-based word composition model. The model is trained in the same manner as done in the \textit{augmented hidden-state + char n-grams} method, i.e., by minimizing the \textsc{mse} between the embeddings that it produces and the task-tuned embeddings of words in the training set (initialized with \textsc{gl}o\textsc{v}e). However, the inputs now are context-aware representations of characters instead of one-hot representations.

\vspace{1 mm}
\noindent
\textbf{Word-sum + char n-grams (\textsc{ws + cng})}, \textbf{Augmented word-sum + char n-grams (\textsc{augmented ws + cng})}, and \textbf{Context word-sum + char n-grams (\textsc{context ws + cng}).} These methods are identical to the \textit{(context/ augmented) hidden-state + char n-grams} methods except that here we include the character n-grams and our character-based word composition model on top of the \textit{word-sum} baseline.

\vspace{1 mm}
\noindent
\textbf{Char hidden-state (\textsc{char hs})} and \textbf{Char word-sum (\textsc{char ws}).} In all the methods described up till now, the input to the core \textsc{rnn} is word embeddings. To gauge whether character-level inputs are themselves sufficient or not, we construct two methods based on the \textit{character to word} (\textsc{c2w}) approach of Ling et al. \shortcite{Ling:15}. For the \textit{char hidden-state} method, the input is one-hot representations of characters from a fixed vocabulary. These representations are encoded into a sequence $w_1$, $\dots$, $w_n$ of intermediate word embeddings by a 2-layer bi-directional \textsc{lstm}. The word embeddings are then fed into a 2-layer \textsc{gru} that transforms them into hidden states $h_1$, $\dots$, $h_n$. Finally, as in the \textit{hidden-state} baseline, an \textsc{lr} layer with softmax activation uses the last hidden state $h_n$ to perform classification while propagating error backwards to train the network. The \textit{char word-sum} method is similar except that once the network has been trained, we use the intermediate word embeddings produced by it to train a \textsc{gbdt} classifier in the same manner as done in the \textit{word-sum} baseline.

\section{Experiments and Results}
\subsection{Experimental setup}
We normalize the input by lowercasing all words and removing stop words. For the \textsc{gru} architecture, we use exactly the same hyper-parameters as Pavlopoulos et al. \shortcite{Pavlopoulos:17},\footnote{The authors have not released their models; we replicate their method based on the details in their paper.} i.e., 128 hidden units, Glorot initialization, cross-entropy loss, and Adam optimizer \cite{adam}. Badjatiya et al. \shortcite{Badjatiya:17} also use the same settings except they have fewer hidden units. The \textsc{lstm} in our character-based word composition model has 256 hidden units while that in our encoder has 64; the \textsc{cnn} has filters of widths varying from 1 to 4. The results we report are with an \textsc{lstm}-based word composition model. In all the models, besides dropout regularization \cite{JMLR:v15:srivastava14a}, we hold out a small part of the training set as validation data to prevent over-fitting. We use 300d embeddings and 1 to 5 character n-grams for Wikipedia and 200d embeddings and 1 to 4 character n-grams for Twitter. We implement the models in \textit{Keras} \cite{keras} with \textit{Theano} back-end. We employ \textit{Lightgbm} \cite{lightgbm} as our \textsc{gdbt} classifier and tune its hyper-parameters using 5-fold grid search.

\subsection{Twitter results}
For the Twitter dataset, unlike previous research \cite{Badjatiya:17,W17-3006}, we report the macro precision, recall, and \textsc{f}$_1$ averaged over 10 folds of stratified \textsc{cv} (Table \ref{twitter}). For a classification problem with $N$ classes, macro precision (similarly, macro recall and macro \textsc{f}$_1$) is given by:
\begin{equation*}
Macro\;P = \frac{1}{N}\sum_{i = 1}^N P_i
\end{equation*}

\noindent where $P_i$ denotes precision on class $i$. Macro metrics provide a better sense of effectiveness on the minority classes \cite{van2013macro}.

We observe that character n-grams (\textsc{cng}) consistently enhance performance, while our augmented approach (\textsc{augmented}) further improves upon the results obtained with character n-grams. All the improvements are statistically significant with $p < 0.05$ under 10-fold \textsc{cv} paired t-test.

As Ling et al. \shortcite{Ling:15} noted in their \textsc{pos} tagging experiments, we observe that the \textsc{char hs} and \textsc{char ws} methods perform worse than their counterparts that use pre-trained word embeddings, i.e., the \textsc{hs} and \textsc{ws} baselines respectively.

To further analyze the performance of our best methods (\textsc{context/augmented ws/hs + cng}), we also examine the results on the racism and sexism classes individually (Table \ref{twitter_neg}). As before, we see that our approach consistently improves over the baselines, and the improvements are statistically significant under paired t-tests.

\begin{table}[ht!]
\centering
\small
\begin{tabular}{| c | c | c | c |}
\hline
\textbf{Method} & \textbf{\textsc{p}} & \textbf{\textsc{r}} & \textbf{\textsc{f}$_1$}\\ \hline
\textsc{hs} & 79.14 & 77.06 & 78.01\\
\textsc{char hs} & 79.48 & 69.00 & 72.36\\
\textsc{hs + cng}$^\dagger$ & 80.36 & \textbf{78.20} & 79.19\\
\textsc{augmented hs + cng}$^\dagger$ & 81.28 & 77.84 & \textbf{79.37}\\
\textsc{context hs + cng}$^\dagger$ & \textbf{81.39} & 77.47 & 79.21\\ \hline
\textsc{ws} & 80.78 & 72.83 & 75.93\\
\textsc{char ws} & 80.04 & 68.17 & 71.94\\
\textsc{ws + cng}$^\dagger$ & 83.16 & 76.60 & 79.31\\
\textsc{augmented ws + cng}$^\dagger$ & \textbf{83.50} & \textbf{77.20} & \textbf{79.80}\\
\textsc{context ws + cng}$^\dagger$ & 83.44 & 77.06 & 79.67\\ \hline
\end{tabular}
\caption{Results on the Twitter dataset. The methods we propose are denoted by $^\dagger$. Our best method (\textsc{augmented ws + cng}) significantly outperforms all other methods.}
\label{twitter}
\end{table}

\begin{table}[ht!]
\centering
\small
\subfloat[Racism]{\begin{tabular}{| c | c | c | c |}
\hline
\textbf{Method} & \textbf{\textsc{p}} & \textbf{\textsc{r}} & \textbf{\textsc{f}$_1$}\\ \hline
\textsc{hs} & 74.15 & 72.46 & 73.24\\
\textsc{augmented hs + cng}$^\dagger$ & 76.28 & \textbf{72.72} & \textbf{74.40}\\
\textsc{context hs + cng}$^\dagger$ & \textbf{76.61} & 72.15 & 74.26\\ \hline
\textsc{ws} & 76.43 & 67.77 & 71.78\\
\textsc{augmented ws + cng}$^\dagger$ & \textbf{78.17} & 72.20 & \textbf{75.01}\\
\textsc{context ws + cng}$^\dagger$ & 77.90 & \textbf{72.26} & 74.91\\ \hline
\end{tabular}}
\qquad
\subfloat[Sexism]{\begin{tabular}{| c | c | c | c |}
\hline
\textbf{Method} & \textbf{\textsc{p}} & \textbf{\textsc{r}} & \textbf{\textsc{f}$_1$}\\ \hline
\textsc{hs} & 76.04 & 68.84 & 72.24\\
\textsc{augmented hs + cng}$^\dagger$ & 80.07 & \textbf{69.28} & \textbf{74.26}\\
\textsc{context hs + cng}$^\dagger$ & \textbf{80.29} & 68.52 & 73.92\\ \hline
\textsc{ws} & 81.75 & 57.37 & 67.38\\
\textsc{augmented ws + cng}$^\dagger$ & 85.61 & \textbf{65.91} & \textbf{74.44}\\
\textsc{context ws + cng}$^\dagger$ & \textbf{85.80} & 65.41 & 74.18\\ \hline
\end{tabular}}
\caption{The baselines (\textsc{ws}, \textsc{hs}) vs. our best approaches ($^\dagger$) on the racism and sexism classes.}
\label{twitter_neg}
\end{table}

\vspace{2 mm}
Additionally, we note that the \textsc{augmented ws + cng} method improves the \textsc{f}$_1$ score of the \textsc{ws + cng} method from 74.12 to 75.01 for the racism class, and from 74.03 to 74.44 for the sexism class. The \textsc{augmented hs + cng} method similarly improves the \textsc{f}$_1$ score of the \textsc{hs + cng} method from 74.00 to 74.40 on the racism class while making no notable difference on the sexism class.

We see that the \textsc{context hs/ws + cng} methods do not perform as well as the \textsc{augmented hs/ws + cng} methods. One reason for this is that the Twitter dataset is not able to expose the methods to enough contexts due to its small size. Moreover, because the collection of this dataset was bootstrapped with a search for certain commonly-used abusive words, many such words are shared across multiple tweets belonging to different classes. Given the above, context-aware character representations perhaps do not provide substantial distinctive information.

\subsection{Wikipedia results}
Following previous work \cite{Pavlopoulos:17,Wulczyn:2017:EMP:3038912.3052591}, we conduct a standard 60:40 train--test split experiment on the two Wikipedia datasets. Specifically, from \textsc{w-tox}, $95,692$ comments ($10.0\%$ abusive) are used for training and $63,994$ ($9.1\%$ abusive) for testing; from \textsc{w-att}, $70,000$ ($11.8\%$ abusive) are used for training and $45,854$ ($11.7\%$ abusive) for testing. Table \ref{wiki} reports the macro \textsc{f}$_1$ scores. We do not report scores from the \textsc{char hs} and \textsc{char ws} methods since they showed poor preliminary results compared to the \textsc{hs} and \textsc{ws} baselines.

\begin{table}[ht!]
\centering
\small
\begin{tabular}{| c | c | c |}
\hline
\textbf{Method} & \textbf{\textsc{w-tox}} & \textbf{\textsc{w-att}} \\ \hline
\textsc{hs} & 88.65 & 86.28\\
\textsc{hs + cng}$^\dagger$ & 89.29 & 87.32\\
\textsc{augmented hs + cng}$^\dagger$ & 89.31 & 87.33\\
\textsc{context hs + cng}$^\dagger$ & \textbf{89.35} & \textbf{87.44}\\ \hline
\textsc{ws} & 85.49 & 84.35\\
\textsc{ws + cng}$^\dagger$ & 87.12 & 85.80\\
\textsc{augmented ws + cng}$^\dagger$ & 87.02 & 85.75\\
\textsc{context ws + cng}$^\dagger$ & \textbf{87.16} & \textbf{85.81}\\ \hline
\end{tabular}
\caption{Macro \textsc{f}$_1$ scores on the two Wikipedia datasets. The current state of the art method for these datasets is \textsc{hs}. $^\dagger$ denotes the methods we propose. Our best method (\textsc{context hs + cng}) outperforms all the other methods.}
\label{wiki}
\end{table}

\vspace{3mm}
Mirroring the analysis carried out for the Twitter dataset, Table \ref{wiki_neg} further compares the performance of our best methods for Wikipedia (\textsc{context/augmented hs + cng}) with that of the state of the art baseline (\textsc{hs}) on specifically the abusive classes of \textsc{w-tox} and \textsc{w-att}.

\begin{table}[ht!]
\centering
\small
\subfloat[\textsc{w-tox}]{\begin{tabular}{| c | c | c | c |}
\hline
\textbf{Method} & \textbf{\textsc{p}} & \textbf{\textsc{r}} & \textbf{\textsc{f}$_1$}\\ \hline
\textsc{hs} & 84.48 & 74.60 & 79.24\\
\textsc{augmented hs + cng}$^\dagger$ & \textbf{85.43} & 76.02 & 80.45\\
\textsc{context hs + cng}$^\dagger$ & 85.42 & \textbf{76.17} & \textbf{80.53}\\ \hline
\end{tabular}}
\qquad
\subfloat[\textsc{w-att}]{\begin{tabular}{| c | c | c | c |}
\hline
\textbf{Method} & \textbf{\textsc{p}} & \textbf{\textsc{r}} & \textbf{\textsc{f}$_1$}\\ \hline
\textsc{hs} & 78.61 & 72.88 & 75.64\\
\textsc{augmented hs + cng}$^\dagger$ & 81.23 & 74.06 & 77.48\\
\textsc{context hs + cng}$^\dagger$ & \textbf{81.39} & \textbf{74.28} & \textbf{77.67}\\ \hline
\end{tabular}}
\caption{The current state of the art baseline (\textsc{hs}) vs. our best methods ($^\dagger$) on the abusive classes of \textsc{w-tox} and \textsc{w-att}.}
\label{wiki_neg}
\end{table}

We observe that the augmented approach substantially improves over the state of the art baseline. Unlike in the case of Twitter, our context-aware setup for word composition is now able to further enhance performance courtesy of the larger size of the datasets which increases the availability of contexts. All improvements are significant ($p < 0.05$) under paired t-tests. We note, however, that the gains we get here with the word composition model are relatively small compared to those we get for Twitter. This difference can be explained by the fact that: (i) Wikipedia comments are less noisy than the tweets and contain fewer obfuscations, and (ii) the Wikipedia datasets, being much larger, expose the methods to more words during training, hence reducing the likelihood of unseen words being important to the semantics of the comments they belong to \cite{Kim:2016:CNL:3016100.3016285}.

Like Pavlopoulos et al. \shortcite{Pavlopoulos:17}, we see that the methods that involve summation of word embeddings (\textsc{ws}) perform significantly worse on the Wikipedia datasets compared to those that use hidden state (\textsc{hs}); however, their performance is comparable or even superior on the Twitter dataset. This contrast is best explained by the observation of Nobata et al. \shortcite{Nobata:2016:ALD:2872427.2883062} that taking average or sum of word embeddings compromises contextual and word order information. While this is beneficial in the case of tweets which are short and loosely-structured, it leads to poor performance of the \textsc{ws} and \textsc{ws + cng} methods on the Wikipedia datasets, with the addition of the word composition model (\textsc{context/augmented ws + cng}) providing little to no improvements.


\begin{table*}[t!]
\centering
\small
\begin{tabular}{| p{8cm} | c | c | c |}
\hline
\hspace{3cm}\textbf{Abusive sample} & \multicolumn{3}{c |}{\textbf{Predicted class}} \\ \hline
 & \textbf{\textsc{ws}} & \textbf{\textsc{ws + cng}} & \textbf{\textsc{augmented ws + cng}} \\ \hline
\textit{@mention I love how the Islamofascists recruit 14 and 15 year old jihadis and then talk about minors in reference to 17 year olds.} & \textit{neither} & \textit{racism} & \textit{racism}\\ \hdashline
\textit{@mention @mention @mention As a certified inmate of the Islamasylum, you don't have the ability to judge.} & \textit{neither} & \textit{racism} & \textit{racism}\\ \hdashline
\textit{@mention ``I'll be ready in 5 minutes" from a girl usually means ``I'll be ready in 20+ minutes." \#notsexist \#knownfromexperience} & \textit{neither} & \textit{sexism} & \textit{sexism}\\ \hdashline
\textit{RT @mention: \#isis \#muslim \#Islamophobia? I think the word you're searching for is \#Islamorealism http://t.co/NyihT8Bqyu http://t.c...} & \textit{neither} & \textit{neither} & \textit{racism}\\ \hdashline
\textit{@mention @mention And looking at your page, I can see that you are in the business of photoshopping images, Islamist cocksucker.} & \textit{neither} & \textit{neither} & \textit{racism}\\ \hdashline
\textit{I think Kat is on the wrong show. \#mkr is for people who can cook. \#stupidbitch \#hopeyouareeliminated} & \textit{neither} & \textit{neither} & \textit{sexism}\\ \hdashline
\textit{@mention because w0m3n are a5sh0les \#feminismishate} & \textit{neither} & \textit{neither} & \textit{sexism}\\ \hline
\end{tabular}
\caption{Improved classification upon the addition of character n-grams (\textsc{cng}) and our word composition model (\textsc{augmented}). Names of users have been replaced with \textit{mention} for anonymity.}
\label{win_examples}
\end{table*}

\section{Discussion}
To investigate the extent to which obfuscated words can be a problem, we extract a number of statistics. Specifically, we notice that out of the approximately $16k$ unique tokens present in the Twitter dataset, there are about $5.2k$ tokens that we cannot find in the English dictionary.\footnote{We use the \textsc{us} \textsc{e}nglish spell-checking utility provided by the \textit{PyEnchant} library of python.} Around $600$ of these $5.2k$ tokens are present in the racist tweets, $1.6k$ in the sexist tweets, and the rest in tweets that are neither. Examples from the racist tweets include \textit{fuckbag}, \textit{ezidiz}, \textit{islamofascists}, \textit{islamistheproblem}, \textit{islamasylum} and \textit{isisaremuslims}, while those from the sexist tweets include \textit{c*nt}, \textit{bbbbitch}, \textit{feminismisawful}, and \textit{stupidbitch}. Given that the racist and sexist tweets come from a small number of unique users, 5 and 527 respectively, we believe that the presence of obfuscated words would be even more pronounced if tweets were procured from more unique users.

In the case of the Wikipedia datasets, around $15k$ unique tokens in the abusive comments of both \textsc{w-tox} and \textsc{w-att} are not attested in the English dictionary. Examples of such tokens from \textsc{w-tox} include \textit{fuggin}, \textit{n*gga}, \textit{fuycker}, and \textit{1d10t}; and from \textsc{w-att} include \textit{f**king}, \textit{beeeitch}, \textit{musulmans}, and \textit{motherfucken}. In comparison to the tweets, the Wikipedia comments use more ``standard'' language. This is validated by the fact that only 14\% of the tokens present in \textsc{w-tox} and \textsc{w-att} are absent from the English dictionary as opposed to 32\% of the tokens in the Twitter dataset even though the Wikipedia ones are almost ten times larger.

Across the three datasets, we note that the addition of character n-gram features enhances the performance of \textsc{rnn}-based methods, corroborating the previous findings that they capture complementary structural and lexical information of words. The inclusion of our character-based word composition model yields state of the art results on all the datasets, demonstrating the benefits of inferring the semantics of unseen words. Table \ref{win_examples} shows some abusive samples from Twitter that are misclassified by the \textsc{ws} baseline method but are correctly classified upon the addition of character n-grams (\textsc{ws + cng}) and the further addition of our character-based word composition model (\textsc{augmented ws + cng}).

Many of the abusive tweets that remain misclassified by the \textsc{augmented ws + cng} method are those that are part of some abusive discourse (e.g., \textit{@Mich\_McConnell Just ``her body" right?}) or contain \textsc{url}s to abusive content (e.g., \textit{@salmonfarmer1: Logic in the world of Islam http://t.co/6nALv2HPc3}).

\begin{table}[t!]
\centering
\small
\begin{tabular}{| c | c |}
\hline
\textbf{Word} & \textbf{Similar words in training set} \\ \hline
\textit{women} & \textit{girls}, \textit{woman}, \textit{females}, \textit{chicks}, \textit{ladies}\\
\textit{w0m3n}$^\dagger$ & \textit{woman}, \textit{women}, \textit{girls}, \textit{ladies}, \textit{chicks}\\
\textit{cunt} & \textit{twat}, \textit{prick}, \textit{faggot}, \textit{slut}, \textit{asshole}\\
\textit{a5sh0les}$^\dagger$ & \textit{assholes}, \textit{stupid}, \textit{cunts}, \textit{twats}, \textit{faggots}\\
\textit{stupidbitch}$^\dagger$ & \textit{idiotic}, \textit{stupid}, \textit{dumb}, \textit{ugly}, \textit{women}\\
\textit{jihad} & \textit{islam}, \textit{muslims}, \textit{sharia}, \textit{terrorist}, \textit{jihadi}\\
\textit{jihaaadi}$^\dagger$ & \textit{terrorists}, \textit{islamist}, \textit{jihadists}, \textit{muslims}\\
\textit{terroristislam}$^\dagger$ & \textit{terrorists}, \textit{muslims}, \textit{attacks}, \textit{extremists}\\
\textit{fuckyouass}$^\dagger$ & \textit{fuck}, \textit{shit}, \textit{fucking}, \textit{damn}, \textit{hell}\\ \hline
\end{tabular}
\caption{Words in the training set that exhibit high cosine similarity to the given word. The ones marked with $^\dagger$ are not seen during training; embeddings for them are composed using our word composition model.}
\label{similar}
\end{table}

In the case of the Wikipedia datasets, there are abusive examples like \textit{smyou have a message re your last change, go fuckyourself!!!} and \textit{F-uc-k you, a-ss-hole Motherf--ucker!} that are misclassified by the state of the art \textsc{hs} baseline and the \textsc{hs + cng} method but correctly classified by our best method for the datasets, i.e., \textsc{context hs + cng}.

To ascertain the effectiveness of our task-tuning process for embeddings, we conducted a qualitative analysis, validating that semantically similar words cluster together in the embedding space. Analogously, we assessed the merits of our word composition model by verifying the neighbors of embeddings formed by it for obfuscated words not seen during training. Table \ref{similar} provides some examples. We see that our model correctly infers the semantics of obfuscated words, even in cases where obfuscation is by concatenation of words.

\section{Conclusions}
In this paper, we considered the problem of obfuscated words in the field of automated abuse detection. Working with three datasets from two different domains, namely Twitter and Wikipedia talk page, we first comprehensively replicated the previous state of the art \textsc{rnn} methods for the datasets. We then showed that character n-grams capture complementary information, and hence, are able to enhance the performance of the \textsc{rnn}s. Finally, we constructed a character-based word composition model in order to infer semantics for unseen words and further extended it with context-aware character representations. The integration of our composition model with the enhanced \textsc{rnn} methods yielded the best results on all three datasets. We have experimentally demonstrated that our approach to modeling obfuscated words significantly advances the state of the art in abuse detection. In the future, we wish to explore its efficacy in tasks such as grammatical error detection and correction. We will make our models and logs of experiments publicly available at \url{https://github.com/pushkarmishra/AbuseDetection}.

\section*{Acknowledgements}
Special thanks to the anonymous reviewers for their valuable comments and suggestions.

\bibliography{emnlp2018}

\begin{thebibliography}{28}
\expandafter\ifx\csname natexlab\endcsname\relax\def\natexlab#1{#1}\fi

\bibitem[{Badjatiya et~al.(2017)Badjatiya, Gupta, Gupta, and
  Varma}]{Badjatiya:17}
Pinkesh Badjatiya, Shashank Gupta, Manish Gupta, and Vasudeva Varma. 2017.
\newblock Deep learning for hate speech detection in tweets.
\newblock In \emph{Proceedings of the 26th International Conference on World
  Wide Web Companion}, WWW '17 Companion, pages 759--760, Republic and Canton
  of Geneva, Switzerland. International World Wide Web Conferences Steering
  Committee.

\bibitem[{Bojanowski et~al.(2017)Bojanowski, Grave, Joulin, and
  Mikolov}]{fasttext}
Piotr Bojanowski, Edouard Grave, Armand Joulin, and Tomas Mikolov. 2017.
\newblock Enriching word vectors with subword information.
\newblock \emph{Transactions of the Association for Computational Linguistics},
  5:135--146.

\bibitem[{Chollet et~al.(2015)}]{keras}
Fran\c{c}ois Chollet et~al. 2015.
\newblock Keras.

\bibitem[{Davidson et~al.(2017)Davidson, Warmsley, Macy, and Weber}]{davidson}
Thomas Davidson, Dana Warmsley, Michael Macy, and Ingmar Weber. 2017.
\newblock Automated hate speech detection and the problem of offensive
  language.
\newblock In \emph{Proceedings of the 11th International AAAI Conference on Web
  and Social Media}, ICWSM '17.

\bibitem[{Djuric et~al.(2015)Djuric, Zhou, Morris, Grbovic, Radosavljevic, and
  Bhamidipati}]{Djuric:2015:HSD:2740908.2742760}
Nemanja Djuric, Jing Zhou, Robin Morris, Mihajlo Grbovic, Vladan Radosavljevic,
  and Narayan Bhamidipati. 2015.
\newblock Hate speech detection with comment embeddings.
\newblock In \emph{Proceedings of the 24th International Conference on World
  Wide Web}, WWW '15 Companion, pages 29--30, New York, NY, USA. ACM.

\bibitem[{Duggan(2014)}]{pew}
Maeve Duggan. 2014.
\newblock Online harassment.

\bibitem[{Forman and Scholz(2010)}]{Forman:10}
George Forman and Martin Scholz. 2010.
\newblock Apples-to-apples in cross-validation studies: Pitfalls in classifier
  performance measurement.
\newblock \emph{SIGKDD Explor. Newsl.}, 12(1):49--57.

\bibitem[{Ke et~al.(2017)Ke, Meng, Finley, Wang, Chen, Ma, Ye, and
  Liu}]{lightgbm}
Guolin Ke, Qi~Meng, Thomas Finley, Taifeng Wang, Wei Chen, Weidong Ma, Qiwei
  Ye, and Tie-Yan Liu. 2017.
\newblock Lightgbm: A highly efficient gradient boosting decision tree.
\newblock In I.~Guyon, U.~V. Luxburg, S.~Bengio, H.~Wallach, R.~Fergus,
  S.~Vishwanathan, and R.~Garnett, editors, \emph{Advances in Neural
  Information Processing Systems 30}, pages 3149--3157. Curran Associates, Inc.

\bibitem[{Kim(2014)}]{D14-1181}
Yoon Kim. 2014.
\newblock Convolutional neural networks for sentence classification.
\newblock In \emph{Proceedings of the 2014 Conference on Empirical Methods in
  Natural Language Processing (EMNLP)}, pages 1746--1751. Association for
  Computational Linguistics.

\bibitem[{Kim et~al.(2016)Kim, Jernite, Sontag, and
  Rush}]{Kim:2016:CNL:3016100.3016285}
Yoon Kim, Yacine Jernite, David Sontag, and Alexander~M. Rush. 2016.
\newblock Character-aware neural language models.
\newblock In \emph{Proceedings of the Thirtieth AAAI Conference on Artificial
  Intelligence}, AAAI'16, pages 2741--2749. AAAI Press.

\bibitem[{Kingma and Ba(2015)}]{adam}
Diederik~P. Kingma and Jimmy Ba. 2015.
\newblock Adam: {A} method for stochastic optimization.
\newblock In \emph{Proceedings of the 3rd International Conference on Learning
  Representations}, ICLR '15.

\bibitem[{Le and Mikolov(2014)}]{DBLP:journals/corr/LeM14}
Quoc Le and Tomas Mikolov. 2014.
\newblock Distributed representations of sentences and documents.
\newblock In \emph{Proceedings of the 31st International Conference on
  International Conference on Machine Learning}, ICML '14.

\bibitem[{Ling et~al.(2015)Ling, Dyer, Black, Trancoso, Fermandez, Amir,
  Marujo, and Luis}]{Ling:15}
Wang Ling, Chris Dyer, Alan~W Black, Isabel Trancoso, Ramon Fermandez, Silvio
  Amir, Luis Marujo, and Tiago Luis. 2015.
\newblock Finding function in form: Compositional character models for open
  vocabulary word representation.
\newblock In \emph{Proceedings of the 2015 Conference on Empirical Methods in
  Natural Language Processing}, pages 1520--1530. Association for Computational
  Linguistics.

\bibitem[{Madhyastha et~al.(2015)Madhyastha, Bansal, Gimpel, and
  Livescu}]{Madhyastha:15}
Pranava~Swaroop Madhyastha, Mohit Bansal, Kevin Gimpel, and Karen Livescu.
  2015.
\newblock Mapping unseen words to task-trained embedding spaces.
\newblock \emph{CoRR}, abs/1510.02387.

\bibitem[{Mishra et~al.(2018)Mishra, Del~Tredici, Yannakoudakis, and
  Shutova}]{mishra}
Pushkar Mishra, Marco Del~Tredici, Helen Yannakoudakis, and Ekaterina Shutova.
  2018.
\newblock Author profiling for abuse detection.
\newblock In \emph{Proceedings of the 27th International Conference on
  Computational Linguistics}, pages 1088--1098. Association for Computational
  Linguistics.

\bibitem[{Nobata et~al.(2016)Nobata, Tetreault, Thomas, Mehdad, and
  Chang}]{Nobata:2016:ALD:2872427.2883062}
Chikashi Nobata, Joel Tetreault, Achint Thomas, Yashar Mehdad, and Yi~Chang.
  2016.
\newblock Abusive language detection in online user content.
\newblock In \emph{Proceedings of the 25th International Conference on World
  Wide Web}, WWW '16, pages 145--153, Republic and Canton of Geneva,
  Switzerland. International World Wide Web Conferences Steering Committee.

\bibitem[{Park and Fung(2017)}]{W17-3006}
Ji~Ho Park and Pascale Fung. 2017.
\newblock One-step and two-step classification for abusive language detection
  on twitter.
\newblock In \emph{Proceedings of the First Workshop on Abusive Language
  Online}, pages 41--45. Association for Computational Linguistics.

\bibitem[{Pavlopoulos et~al.(2017)Pavlopoulos, Malakasiotis, and
  Androutsopoulos}]{Pavlopoulos:17}
John Pavlopoulos, Prodromos Malakasiotis, and Ion Androutsopoulos. 2017.
\newblock Deep learning for user comment moderation.
\newblock In \emph{Proceedings of the First Workshop on Abusive Language
  Online}, pages 25--35. Association for Computational Linguistics.

\bibitem[{Pennington et~al.(2014)Pennington, Socher, and
  Manning}]{Pennington:14}
Jeffrey Pennington, Richard Socher, and Christopher~D. Manning. 2014.
\newblock Glove: Global vectors for word representation.
\newblock In \emph{Empirical Methods in Natural Language Processing (EMNLP)},
  pages 1532--1543.

\bibitem[{Pinter et~al.(2017)Pinter, Guthrie, and Eisenstein}]{pinter}
Yuval Pinter, Robert Guthrie, and Jacob Eisenstein. 2017.
\newblock Mimicking word embeddings using subword rnns.
\newblock In \emph{Proceedings of the 2017 Conference on Empirical Methods in
  Natural Language Processing}, pages 102--112. Association for Computational
  Linguistics.

\bibitem[{Qian et~al.(2018)Qian, ElSherief, Belding, and Wang}]{leveraging}
J.~Qian, M.~ElSherief, E.~Belding, and W.~Wang. 2018.
\newblock Leveraging intra-user and inter-user representation learning for
  automated hate speech detection.
\newblock \emph{NAACL HLT, New Orleans, LA, June 2018.}, page \textit{to
  appear}.

\bibitem[{Srivastava et~al.(2014)Srivastava, Hinton, Krizhevsky, Sutskever, and
  Salakhutdinov}]{JMLR:v15:srivastava14a}
Nitish Srivastava, Geoffrey Hinton, Alex Krizhevsky, Ilya Sutskever, and Ruslan
  Salakhutdinov. 2014.
\newblock Dropout: A simple way to prevent neural networks from overfitting.
\newblock \emph{Journal of Machine Learning Research}, 15:1929--1958.

\bibitem[{Van~Asch(2013)}]{van2013macro}
Vincent Van~Asch. 2013.
\newblock Macro-and micro-averaged evaluation measures [[basic draft]].
\newblock \emph{Computational Linguistics \& Psycholinguistics, University of
  Antwerp, Belgium}.

\bibitem[{Waseem(2016)}]{zeerakW16-5618}
Zeerak Waseem. 2016.
\newblock Are you a racist or am {I} seeing things? annotator influence on hate
  speech detection on twitter.
\newblock In \emph{Proceedings of the First Workshop on NLP and Computational
  Social Science}, pages 138--142. Association for Computational Linguistics.

\bibitem[{Waseem et~al.(2017)Waseem, Chung, Hovy, and
  Tetreault}]{waseem2017proceedings}
Zeerak Waseem, Wendy Hui~Kyong Chung, Dirk Hovy, and Joel Tetreault. 2017.
\newblock Proceedings of the first workshop on abusive language online.
\newblock In \emph{Proceedings of the First Workshop on Abusive Language
  Online}. Association for Computational Linguistics.

\bibitem[{Waseem and Hovy(2016)}]{c53cecce142c48628b3883d13155261c}
Zeerak Waseem and Dirk Hovy. 2016.
\newblock Hateful symbols or hateful people? predictive features for hate
  speech detection on twitter.
\newblock In \emph{Proceedings of the NAACL Student Research Workshop}, pages
  88--93, San Diego, California. Association for Computational Linguistics.

\bibitem[{Wulczyn et~al.(2017)Wulczyn, Thain, and
  Dixon}]{Wulczyn:2017:EMP:3038912.3052591}
Ellery Wulczyn, Nithum Thain, and Lucas Dixon. 2017.
\newblock Ex machina: Personal attacks seen at scale.
\newblock In \emph{Proceedings of the 26th International Conference on World
  Wide Web}, WWW '17, pages 1391--1399, Republic and Canton of Geneva,
  Switzerland. International World Wide Web Conferences Steering Committee.

\bibitem[{Yin et~al.(2009)Yin, Davison, Xue, Hong, Kontostathis, and
  Edwards}]{Yin09detectionof}
Dawei Yin, Brian~D. Davison, Zhenzhen Xue, Liangjie Hong, April Kontostathis,
  and Lynne Edwards. 2009.
\newblock Detection of harassment on web 2.0.
\newblock In \emph{Processings of the Content Analysis in the WEB 2.0}, 2:1-7.

\end{thebibliography}
\bibliographystyle{emnlp_natbib_nourl}
\end{document}